\documentclass[final]{cvpr}

\usepackage{times}
\usepackage{epsfig}
\usepackage{graphicx}
\usepackage{amsmath}
\usepackage{amssymb}

\usepackage{booktabs}
\usepackage[caption=false]{subfig}

\usepackage[pagebackref=true,breaklinks=true,colorlinks,bookmarks=false]{hyperref}

\def\cvprPaperID{****} %
\def\confYear{CVPR 2021}

\begin{document}

\title{NTIRE 2021 Challenge on Burst Super-Resolution:
Methods and Results}

\author{Goutam Bhat$^*$ \and Martin Danelljan$^*$ \and Radu Timofte$^*$ \and Kazutoshi Akita \and Wooyeong Cho \and Haoqiang Fan \and Lanpeng Jia \and Daeshik Kim \and Bruno Lecouat \and Youwei Li \and
Shuaicheng Liu \and Ziluan Liu \and Ziwei Luo \and Takahiro Maeda \and Julien Mairal \and Christian Micheloni \and Xuan Mo \and Takeru Oba 
\and Pavel Ostyakov \and Jean Ponce \and Sanghyeok Son \and 
Jian Sun \and Norimichi Ukita \and Rao Muhammad Umer \and Youliang Yan \and Lei Yu \and Magauiya Zhussip \and Xueyi Zou
}

\maketitle

\begin{abstract}
This paper reviews the NTIRE2021 challenge on burst super-resolution. Given a RAW noisy burst as input, the task in the challenge was to generate a clean RGB image with 4 times higher resolution. The challenge contained two tracks; Track 1 evaluating on synthetically generated data, and Track 2 using real-world bursts from mobile camera. In the final testing phase, 6 teams submitted results using a diverse set of solutions. The top-performing methods set a new state-of-the-art for the burst super-resolution task.
\end{abstract}

{\let\thefootnote\relax\footnote{{$^*$Goutam Bhat, Martin Danelljan, and Radu Timofte at ETH Zurich are the NTIRE 2021 challenge organizers. The other authors participated in the challenge and are listed alphabetically. \\
Appendix A contains the participants' team names and affiliations. \\
NTIRE 2021 webpage:\\ \url{https://data.vision.ee.ethz.ch/cvl/ntire21/}}}}

\newcommand{\parsection}[1]{\vspace{0.5mm}\noindent\textbf{#1:}~}
\newcommand{\MD}[1]{\textcolor{purple}{\textbf{[MD:} \textit{#1}\textbf{]}}}
\newcommand{\GB}[1]{\textcolor{blue}{\textbf{[GB:} \textit{#1}\textbf{]}}}

\section{Introduction}
Super-resolution (SR) is a fundamental computer vision problem with numerous applications in \eg mobile photography, remote sensing, medical imaging. Given a single or multiple images of a scene, SR aims to generate a higher-resolution output by adding missing high-frequency details. In recent years, the SR community has mainly focused on the single-image super-resolution (SISR) task~\cite{Dong2014LearningAD,Dong2016ImageSU,Lai2017DeepLP,Lim2017EnhancedDR,Kim2016AccurateIS,Zhang2018ResidualDN,Tai2017ImageSV,Johnson2016PerceptualLF,Ledig2017PhotoRealisticSI,Yu2016UltraResolvingFI,Wang2018ESRGANES,lugmayr2020srflow}. Thanks to the development of specialized network architectures~\cite{Dong2016ImageSU,Lai2017DeepLP,Lim2017EnhancedDR,Kim2016AccurateIS,Zhang2018ResidualDN,Tai2017ImageSV} and training strategies~\cite{Johnson2016PerceptualLF,Ledig2017PhotoRealisticSI,Yu2016UltraResolvingFI,Wang2018ESRGANES}, the SISR methods have achieved impressive SR performance. Despite these advances, the SISR approaches are fundamentally limited by the available information (single frame), and thus only rely on learned image priors to recover the missing details.

In contrast, multi-frame super-resolution (MFSR) approaches combine information from multiple low-resolution (LR) images to generate a HR output. If the input images are captured using a non-stationary camera and thus contain sub-pixel shifts \wrt each other, they provide multiple LR samplings of the same underlying scene. By effectively fusing this information, the MFSR methods can reconstruct high-frequency details which otherwise cannot be recovered using a single input image. This makes MFSR especially tempting for the popular mobile burst photography applications. Since the burst images contain sub-pixel shifts due to natural hand tremors~\cite{Wronski2019HandheldMS}, MFSR can be employed to improve the image resolution which is otherwise restricted by hardware constraints. 

Despite the aforementioned advantages, the MFSR problem has received limited attention in recent years, compared to the SISR task. The recent work~\cite{Bhat2021DeepBS} by Bhat~\etal aims to address this issue by introducing a synthetic, as well as a real-world dataset for burst super-resolution, in addition to a MFSR network architecture. The NTIRE 2021 Challenge on Burst Super-Resolution aims to further stimulate research in the burst super-resolution task. The challenge consists of two tracks. In Track 1, the methods are evaluated on the synthetic burst dataset introduced in~\cite{Bhat2021DeepBS}, and ranked using standard fidelity score PSNR. Track 2 evaluates the real world performance on the BurstSR dataset introduced in~\cite{Bhat2021DeepBS}. The BurstSR dataset consists of bursts captured using a hand held camera, along with a corresponding HR image captured using a DSLR. The methods are ranked using a combination of fidelity score as well as a human study.

In total, 6 teams participated in the NTIRE 2021 Challenge on Burst Super-Resolution. The participating teams employed a variety of fusion approaches, alignment modules, and reconstruction networks. 4 of the 6 participating teams outperformed DBSR~\cite{Bhat2021DeepBS} on Track 1, setting a new state-of-the-art on the burst super-resolution task.

\section{NTIRE 2021 Challenge}
The goal of the NTIRE 2021 Challenge on Burst Super-Resolution is to encourage further research in the burst SR task and provide a common benchmark for evaluating different methods. This challenge is one of the NTIRE 2021 associated challenges: nonhomogeneous dehazing~\cite{ancuti2021ntire}, defocus deblurring using dual-pixel~\cite{abuolaim2021ntire}, depth guided image relighting~\cite{elhelou2021ntire}, image deblurring~\cite{nah2021ntire}, multi-modal aerial view imagery classification~\cite{liu2021ntire}, learning the super-resolution space~\cite{lugmayr2021ntire}, quality enhancement of heavily compressed videos~\cite{yang2021ntire}, video super-resolution~\cite{son2021ntire}, perceptual image quality assessment~\cite{gu2021ntire}, burst super-resolution, and high dynamic range~\cite{perez2021ntire}.
The burst super-resolution challenge contained two tracks. In both tracks, the methods are provided a noisy RAW burst containing 14 images. The task is to perform joint denoising, demosaicking, and super-resolution to generate a clean RGB image with 4 times higher resolution. The participants were provided a public toolkit (\url{https://github.com/goutamgmb/NTIRE21_BURSTSR}) containing tools for training and evaluation for both tracks.
Next, we describe the two challenge tracks in more detail.

\subsection{Track 1: Synthetic}
Track 1 employs synthetic bursts generated using the data generation pipeline employed in~\cite{Bhat2021DeepBS}. Given an sRGB image, an inverse camera pipeline introduced in~\cite{Brooks2019UnprocessingIF} is employed to convert the sRGB image to linear sensor space. Next, a synthetic burst is generated by applying random translations from the range $[-24, 24]$ pixels, and random rotations from the range $[-1, 1]$ degrees. Each image in the burst is then downsampled by a factor of 4 using bilinear interpolation and then mosaicked using Bayer pattern. Finally, independent read and shot noise is added to each image to obtain the noisy RAW burst.  Due to the use of synthetically generated data, an accurately aligned ground truth image is readily available in Track 1. This allows evaluating the impact of different architectural choices and loss functions on the SR performance, computed in terms of pixel-wise image quality metrics such as PSNR.

The public toolkit provided to the participants contained data generation scripts which could be used to generate synthetic bursts for training.
We used the pre-generated synthetic burst dataset introduced in~\cite{Bhat2021DeepBS} as the validation set for Track 1. The dataset consists of 300 bursts, which have been generated using sRGB images from the Zurich RAW to RGB~\cite{ignatov2020replacing} test set. Each burst contains 14 RAW images of resolution $96 \times 96$. The participants could evaluate their methods on the validation set using an evaluation server during the development phase of the challenge. A public leaderboard (\url{https://competitions.codalab.org/competitions/28078#results}) was also made available. The dataset for the final test phase, consisting of 500 bursts, was synthetically generated using the DSLR images from the BurstSR~\cite{Bhat2021DeepBS} test set. Similar to the validation set, each burst in the test set contains 14 $96 \times 96$ RAW images. The participants were only provided the RAW LR bursts, and asked to submit the predictions of their methods.

\subsection{Track 2: Real-World}
In this track, we employ the BurstSR dataset introduced in~\cite{Bhat2021DeepBS} for evaluating the methods. The BurstSR dataset consists of 200 RAW bursts captured from a hand held mobile camera. A corresponding higher-resolution image captured using a DSLR is also provided for each burst to serve as the ground truth. Compared to the synthetic dataset used in Track 1, the BurstSR dataset allows evaluating the performance of the methods on real-world degradation and noise. However, as the input burst and the HR ground truth are captured using different cameras, there are spatial mis-alignment and color differences between the two. This poses additional challenges on the training of the methods as well as evaluation. 

The BurstSR dataset is split into train, validation, and test splits consisting of 160, 20, and 20 bursts, respectively. We extracted $160 \times 160$ crops from the bursts to obtain our training, validation, and test sets consisting of 5405, 882, and 639 crops, respectively.
The participants were allowed to use the provided training set, in addition to external synthetic data, for training their methods. During the development phase, the participants were also provided the validation set, along with the ground truth, for evaluating different design choices. 
Unlike in Track 1, there was no evaluation server in Track 2. For the final test phase, the participants were provided only the LR bursts from the test set, and asked to submit the network predictions.

\section{Challenge Results}
In this section, we report the final results on the test sets of both Track 1 and Track 2. During the final test phase, the participants were asked to submit their predictions on the provided test data. In Track 1, there were 6 teams which submitted their methods, while 5 different teams submitted methods in Track 2. All the submitted methods are briefly described in Section~\ref{sec:team_desc}, while the members and affiliations for each team are listed in Appendix A.

\subsection{Evaluation Metrics}
\label{sec:eval_metrics}
The aim in MFSR is to reconstruct the original HR image by fusing information from multiple LR observations. Thus, we employ fidelity based image metrics to evaluate the prediction quality for different methods. Since there are spatial and color mis-alignments between the input bursts and HR ground truth in the BurstSR dataset employed for Track 2, we additionally conducted a human study to evaluate the top ranking methods.

\parsection{Track 1} Due to the use of synthetically generated dataset for evaluation, an accurately aligned ground truth is available in Track 1. This enables the use of pixel-wise image quality metrics for evaluating the performance of different methods. We use the fidelity-based Peak Signal-to-Noise Ratio (PSNR) score to rank the methods. Additionally, we also report the Structural Similarity Index (SSIM)~\cite{Wang2004ImageQA} as well as the learned perceptual score LPIPS~\cite{Zhang2018TheUE} for all the methods. The emphasis of the challenge is on learning to recover the HR signal, rather than learning any post-processing steps. Thus, all metrics are computed in the linear sensor space, before applying white-balancing, gamma correction, contrast enhancement \etc. 

\parsection{Track 2} We follow the same evaluation procedure employed in~\cite{Bhat2021DeepBS} in order to handle the spatial and color mis-alignments between the input bursts and HR ground truth. The network prediction is first spatially aligned to the ground truth, using pixel-wise optical flow estimated using PWC-Net~\cite{Sun2018PWCNetCF}. A linear color mapping between the input burst and the ground truth, modeled as a 3x3 color correction matrix, is then estimated and used to transform the spatially aligned network prediction to the same color space as the ground truth. The spatially aligned and color corrected prediction is then compared with the ground truth to compute standard image quality metrics. The evaluation script was included in the public toolkit released to the participants. We refer to~\cite{Bhat2021DeepBS} for more details about the evaluation procedure. 

We also conducted a human study on Amazon Mechanical Turk (AMT) to evaluate the top performing methods on Track 2. We manually selected $100$ bursts from the test set with diverse texture.  Next, we extract 3 random $80 \times 80$ crops from the HR predictions of the methods for each of these test images. The crops are then resized to $320 \times 320$ using nearest neighbor interpolation. The participants in the user study were shown the full ground truth image, as well as the network prediction crops, and then asked to rank the predictions based on visual quality. We obtained 5 independent rankings for each crop. The mean ranking (MOR) over the 300 crops, as well as the percentage of times a method was ranked first ($\%$Top) are used as evaluation scores to rank the methods.   

\subsection{Baselines}
We compare the participating methods with two additional baselines.

\parsection{SingleImage} We evaluate a baseline single frame SR method. Our single image baseline passes the first image in the burst through a series of residual blocks~\cite{He2016DeepRL} without batch normalization~\cite{DBLP:conf/icml/IoffeS15}. The extracted feature map is upsampled using the sub-pixel convolution layer~\cite{Shi2016RealTimeSI}, and passed through additional residual blocks to obtain the HR RGB image. 

\parsection{DBSR~\cite{Bhat2021DeepBS}} We also evaluate the DBSR burst super-resolution model introduced in~\cite{Bhat2021DeepBS}. DBSR employs the PWCNet~\cite{Sun2018PWCNetCF} optical flow network align the input images. The aligned images are merged using an attention-based fusion approach.

\subsection{Track 1: Synthetic}
\begin{table}[!t]
	\centering\vspace{-1mm}
	\resizebox{\columnwidth}{!}{%
		\begin{tabular}{lccc}
\toprule
&PSNR$\uparrow$&SSIM$\uparrow$&LPIPS$\downarrow$\\\midrule
Noah\_TerminalVision\_SR & \textbf{46.85} &  \textbf{0.983} & \textbf{0.018} \\
MegSR & 46.72 & \textbf{0.983} & 0.020 \\
Inria & 44.76 & 0.969 & 0.034 \\
TTI &  44.40 &  0.973 & 0.038 \\
BREIL  & 39.22  & 0.918 & 0.104 \\
MLP\_BSR &  37.62  & 0.895  & 0.166 \\\hline
SingleImage & 39.28 & 0.921 & 0.103 \\
DBSR~\cite{Bhat2021DeepBS}& 42.58 & 0.960 & 0.055 \\\bottomrule
\end{tabular}

	}\vspace{1mm}%
	\caption{Challenge results on the synthetic test set from Track 1, in terms of PSNR, SSIM, and LPIPS. The top section contains results for the participating methods, while baseline approaches are included in the bottom section.}
	\label{tab:res_track1}%
	\vspace{0mm}
\end{table}

\begin{figure*}[ht]
    \centering%
    \includegraphics[trim = 0 0 0 0, width=0.95\textwidth]{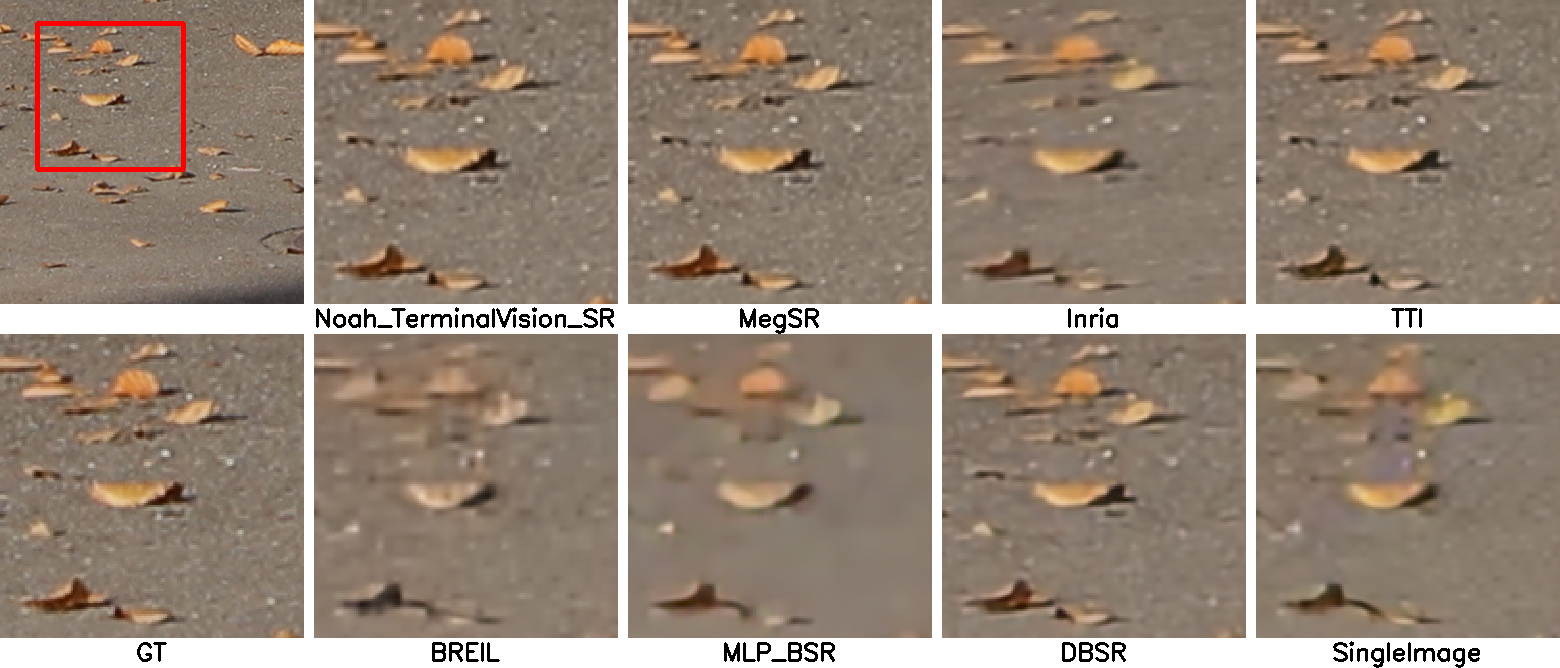}\vspace{1mm} \\
    \includegraphics[trim = 0 0 0 0, width=0.95\textwidth]{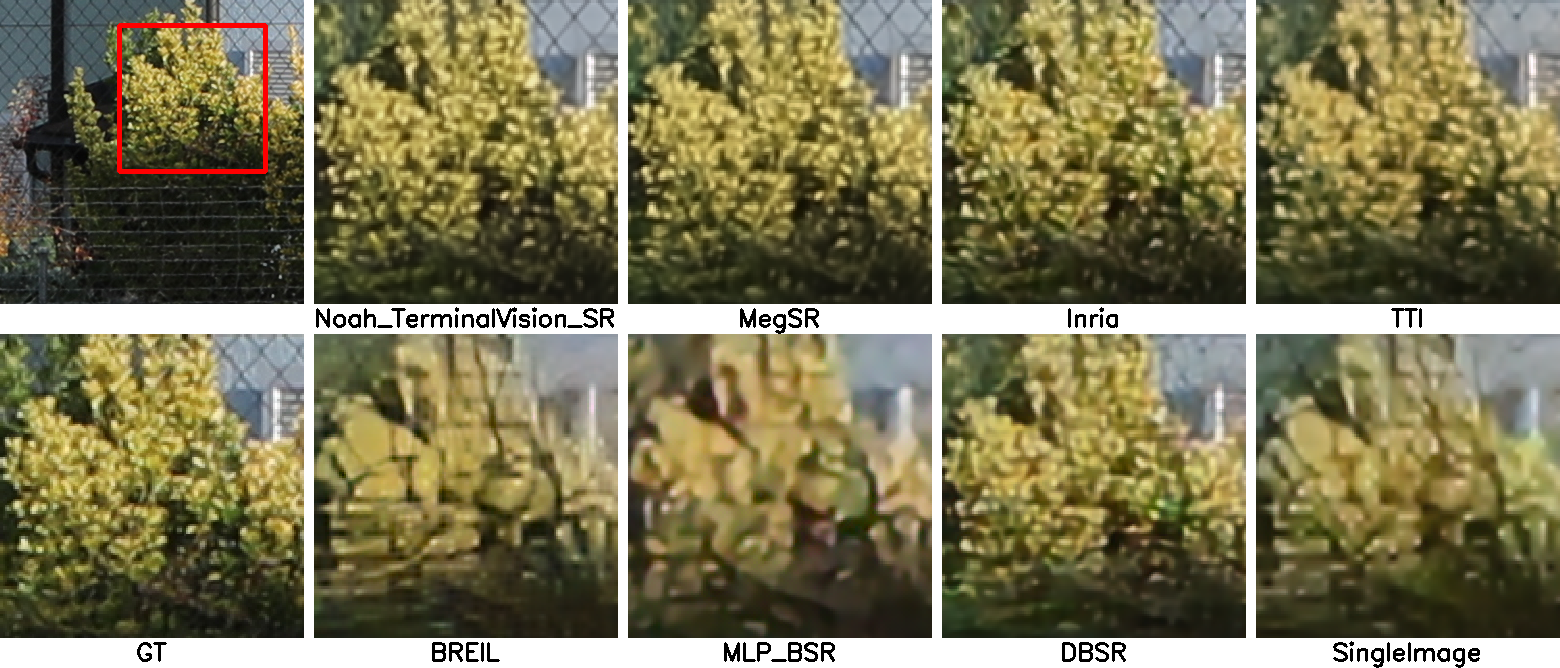}\vspace{1mm} \\
    \includegraphics[trim = 0 0 0 0, width=0.95\textwidth]{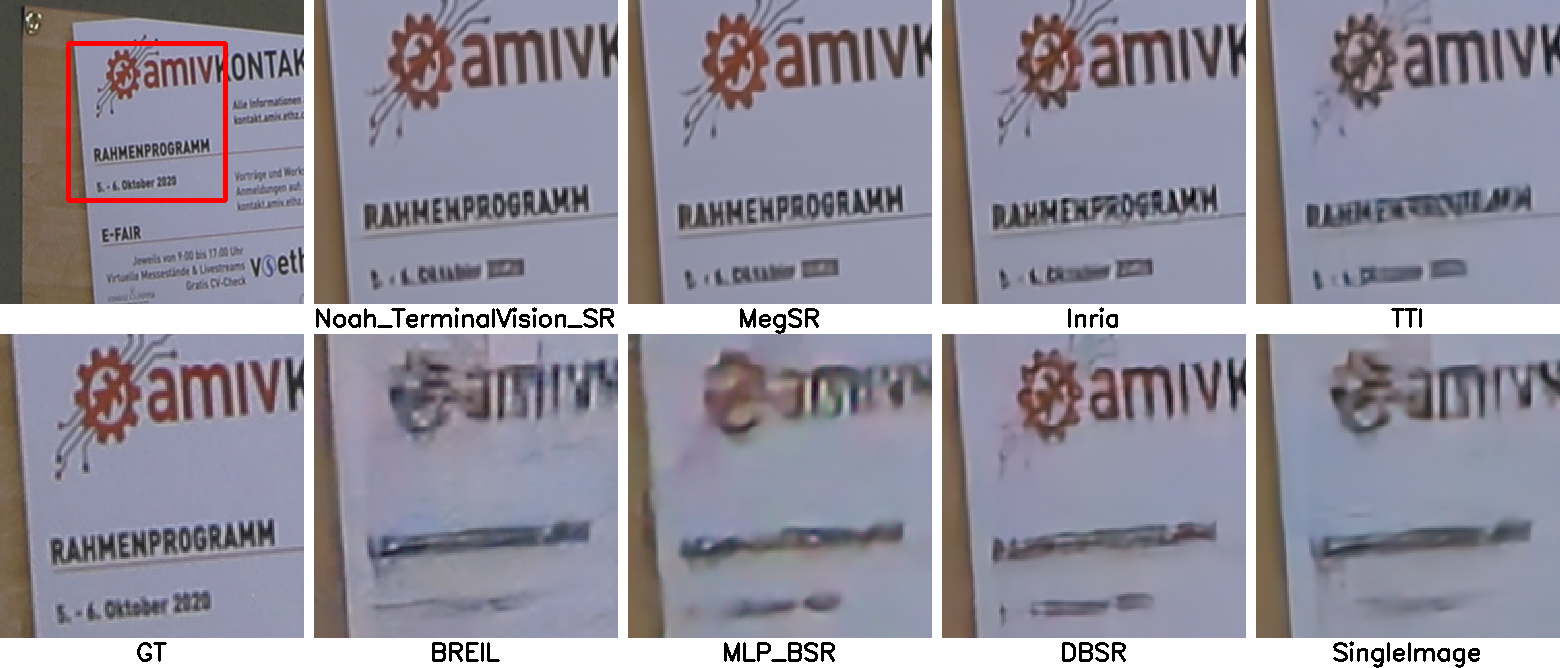}\vspace{0mm}
    \caption{Qualitative comparison on Track 1 test set (4x super-resolution).}\vspace{0mm}
    \label{fig:qual_fig_track1}
\end{figure*}

In this section, we present results of the submitted methods on Track 1. The mean PSNR, SSIM, and LPIPS scores over the 500 bursts from the synthetic test set are provided in Table~\ref{tab:res_track1}. The team Noah\_TerminalVision\_SR obtains the best results in terms of all 3 metrics with a PSNR score of $46.85$. Noah\_TerminalVision\_SR employs the PCD module~\cite{wang2019edvr} to align the input images, which are then merged using the attention-based fusion approach proposed in~\cite{Bhat2021DeepBS}. MegSR achieves the second best performance with a PSNR of $46.72$. MegSR uses a modified version of PCD, denoted FEPCD module for alignment. Fusion is performed using a cross non-local fusion module based on Non-Local network~\cite{Wang_2018_CVPR}. Team Inria, employing an optimization  based approach, obtains a PSNR score of $44.76$. The team utilizes the forward image formation model and jointly optimizes the HR estimate as well as the motion vectors. Note that four of the participating teams, namely Noah\_TerminalVision\_SR, MegSR, Inria, and TTI outperformed the DBSR method~\cite{Bhat2021DeepBS}, thus setting a new state-of-the-art on the burst SR problem. 

A qualitative comparison between the participating methods is provided in Figure~\ref{fig:qual_fig_track1}. The two top performing methods Noah\_TerminalVision\_SR and MegSR obtain impressive results which are very close to the ground truth. The multi-frame approaches Noah\_TerminalVision\_SR, MegSR, Inria, TTI, and DBSR better recover the high-frequency details compared to the single image baseline, thanks to the use of additional information from multiple frames.

\subsection{Track 2: Real-World}

\begin{table}[!t]
	\centering\vspace{-1mm}
	\resizebox{\columnwidth}{!}{%
		\begin{tabular}{lcccccc}
\toprule
&PSNR$\uparrow$&SSIM$\uparrow$&LPIPS$\downarrow$&MOR$\downarrow$&\%\text{Top} $\uparrow$& Avg. Rank$\downarrow$\\\midrule
MegSR & \textbf{45.45} & \textbf{0.979} & 0.032 & 3.09 & 20.9 & \textbf{1.5}\\
Noah\_TerminalVision\_SR\_B & 45.26 & 0.978 & \textbf{0.026} & \textbf{2.96} & \textbf{29.7} & 2.0 \\
Noah\_TerminalVision\_SR\_A & 45.36 & \textbf{0.979} & 0.035 & 3.41 & 13.1 & 2.5 \\
TTI &  44.16 &  0.974 & 0.040 & 3.87 & 14.0 & 5.0\\
MLP\_BSR &  41.40 & 0.952 & 0.101 & - & - & -\\
BREIL  & 29.93  & 0.797 & 0.141 & - & - & -\\\hline
DBSR~\cite{Bhat2021DeepBS}&  45.17 & 0.978 & 0.037 & 3.57 & 11.4 & 4.0\\
SingleImage & 44.02 & 0.972 & 0.051 & 4.10 & 10.9 & 6.0 \\\bottomrule
\end{tabular}

	}\vspace{1mm}%
	\caption{Challenge results on BurstSR test set from Track 2. PSNR, SSIM, and LPIPS scores are computed after spatial and color alignment of the network prediction to the ground truth. MOR denotes the mean ranking of the method from a human study, while $\%$Top corresponds to the percentage of time a method was ranked as the best in the human study. The last column reports the average of a methods ranking in terms of PSNR and MOR.}
	\label{tab:res_track2}%
	\vspace{-3mm}
\end{table}

Here, we present the results on the real-world BurstSR test set from Track 2. The mean PSNR, SSIM, and LPIPS scores over the test set are provided in Table~\ref{tab:res_track2}. Note that all the metrics are computed after spatial and color alignment of the network prediction to the ground truth, as described in Section~\ref{sec:eval_metrics}. MegSR achieves the best results in terms of PSNR and SSIM, with scores of $45.45$dB and 0.979, respectively. Noah\_TerminalVision\_SR\_A, which is trained using $L_1$ loss with alignment similar to MegSR, obtains the second best PSNR score of $45.36$dB. In contrast, Noah\_TerminalVision\_SR\_B which is trained using perceptual loss achieves the best LPIPS score of $0.026$.

\begin{figure*}[ht]
    \centering%
    \includegraphics[trim = 0 0 0 0, width=0.95\textwidth]{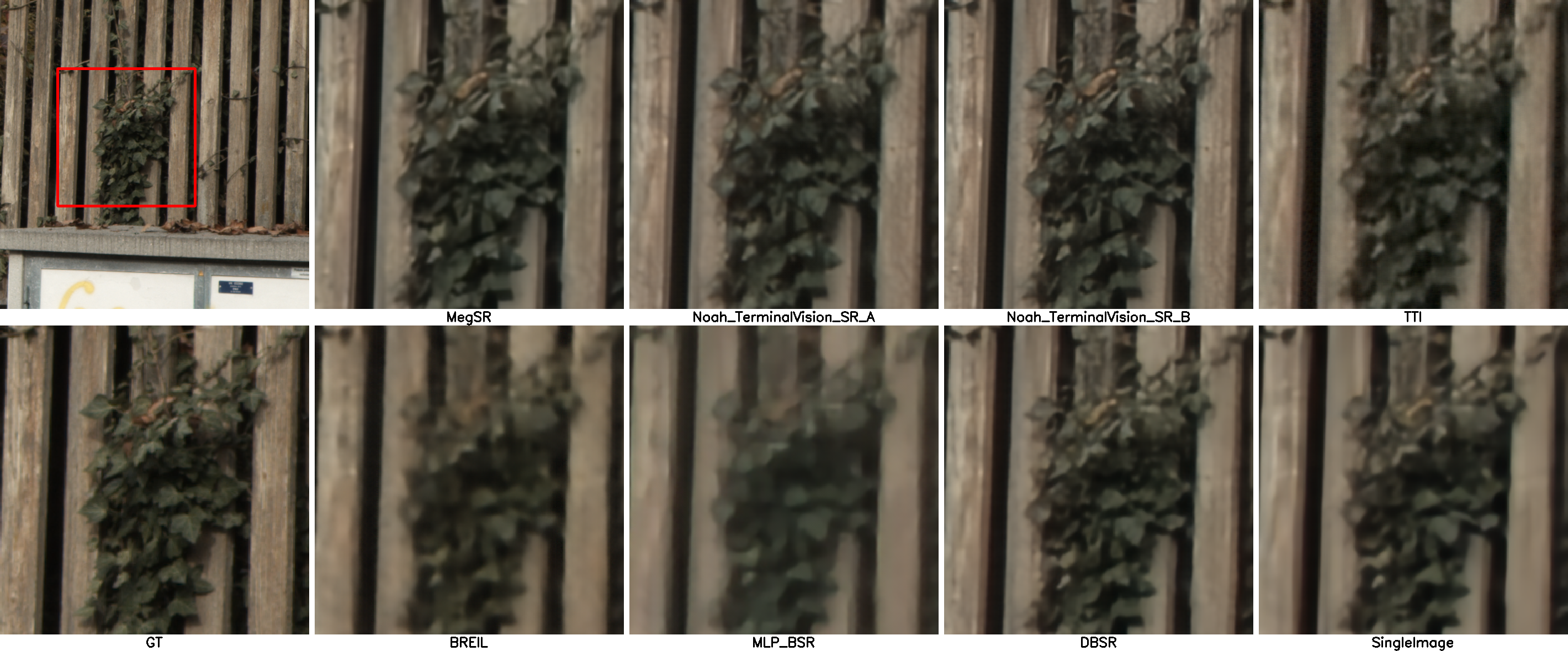}\vspace{1mm} \\
    \includegraphics[trim = 0 0 0 0, width=0.95\textwidth]{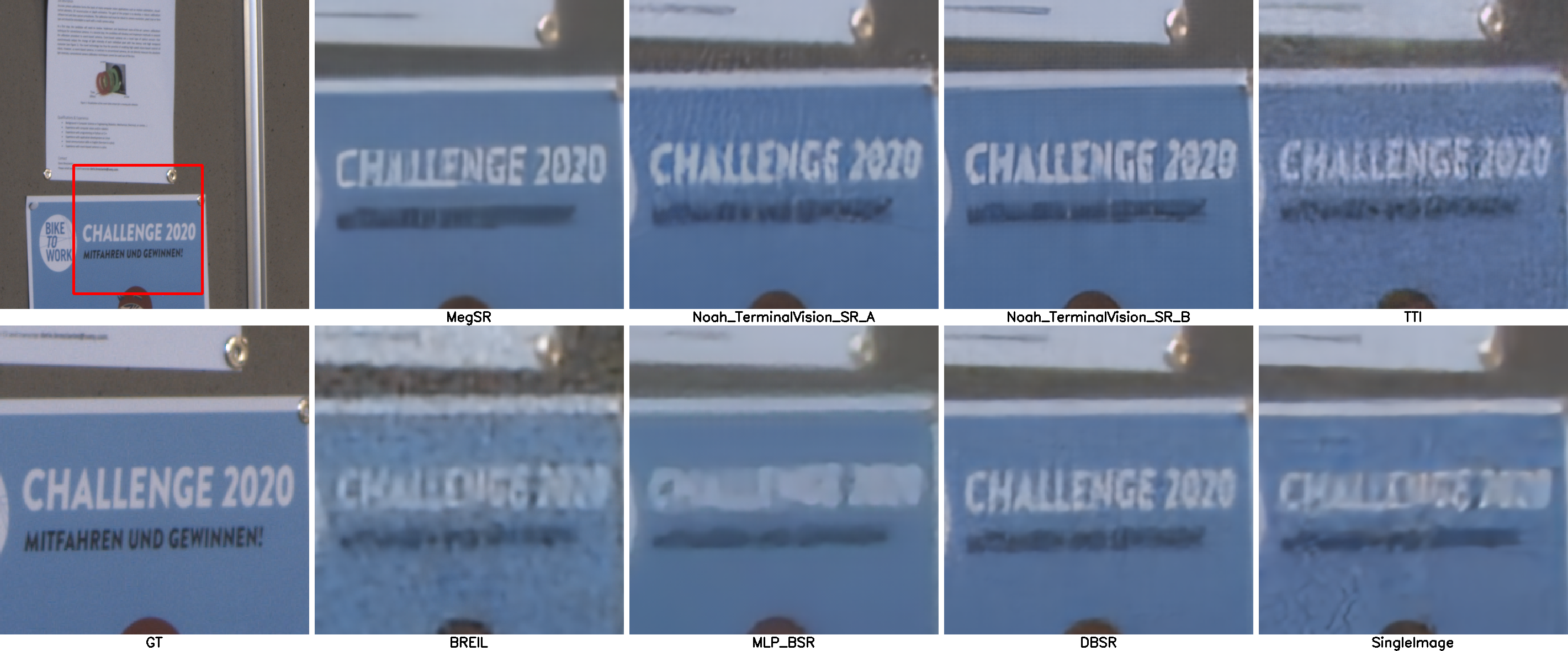}\vspace{1mm} \\
    \includegraphics[trim = 0 0 0 0, width=0.95\textwidth]{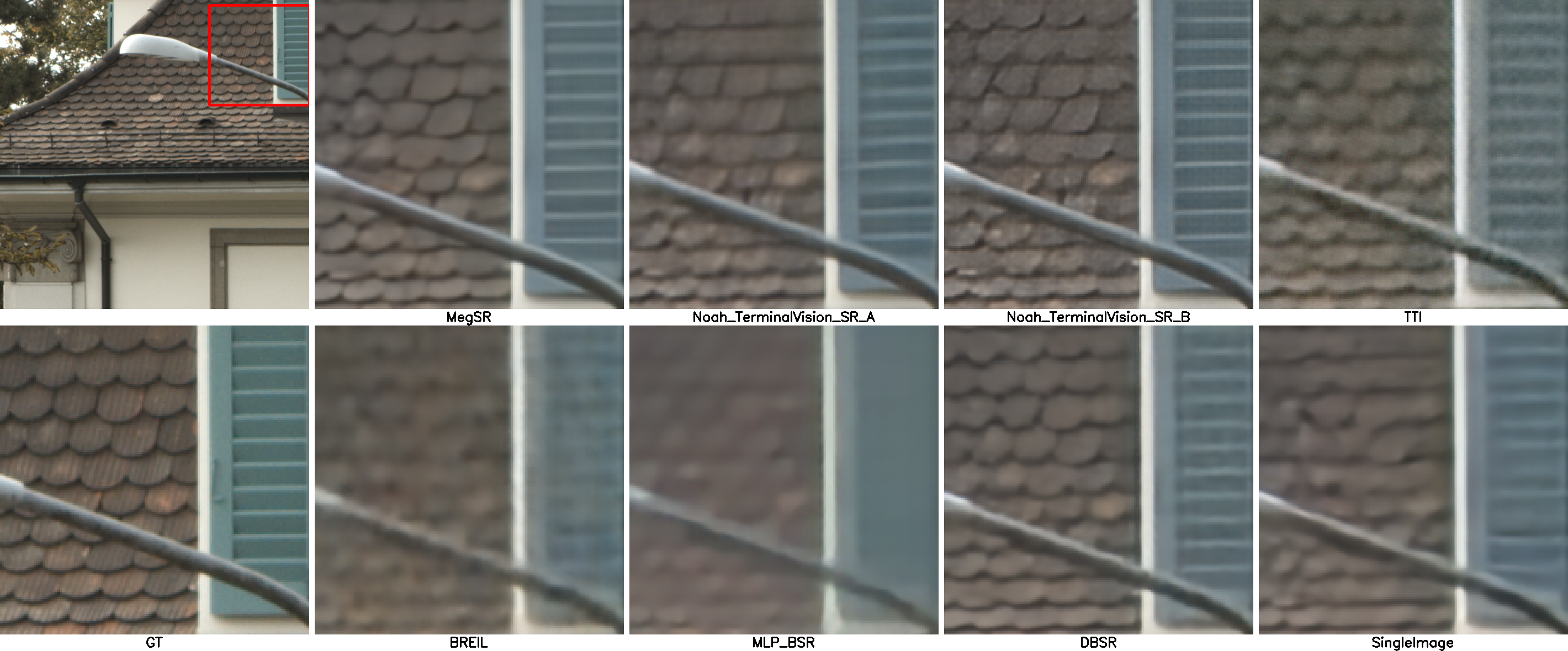}\vspace{0mm}
    \caption{Qualitative comparison on Track 2 test set (4x super-resolution).}\vspace{0mm}
    \label{fig:qual_fig_track2}
\end{figure*}

We also report results of the human study in Table~\ref{tab:res_track2}, in terms of Mean Opinion Ranking (MOR) and $\%$Top metrics. Noah\_TerminalVision\_SR\_B obtains the best MOR score of 2.96, while the second best results are obtained by MegSR. A qualitative comparison between the methods is provided in Figure~\ref{fig:qual_fig_track2}. As in Track 1, MegSR, Noah\_TerminalVision\_SR, TTI, and DBSR generate more detailed images compared to the single image baseline. The results of Noah\_TerminalVision\_SR\_B are in general more sharper compared to that of MegSR. This can be attributed to the use of perceptual loss during training. However the predictions of Noah\_TerminalVision\_SR\_B contain slight high-frequency artefacts, as seen in third example in Figure~\ref{fig:qual_fig_track2}.

\section{Challenge Methods and Teams}
\label{sec:team_desc}
\begin{table*}[!t]
	\centering\vspace{-1mm}
	\resizebox{\textwidth}{!}{%
		\begin{tabular}{lccccc}
\toprule
& & \multicolumn{2}{c}{Track 1}& \multicolumn{2}{c}{Track 2} \\
Team Name&Codalab Username& Train Time (days) & Runtime (sec)& Train Time (days) & Runtime (sec)\\\midrule
Noah\_TerminalVision\_SR &  Noah\_TerminalVision & 5 & 1.60 & 2 & 0.30\\
MegSR &  megviiLuo & 7 & 0.15 & 3 & 0.18\\
Inria &  JohnDoe4598 & 2.5 & 0.65 & - & -\\
TTI & TakahiroMaeda & 18 & 2.14 & 16 & 5.50 \\
BREIL  &  chowy333 & 2.5 & 0.01 & 0.1 & 0.01\\
MLP\_BSR & raoumer & 10 & 0.33 & - & 0.88  \\\bottomrule
\end{tabular}

	}\vspace{1mm}%
	\caption{Information about the participating teams. We report the training time and inference time per burst provided by the teams. }
	\label{tab:team_info}%
	\vspace{0mm}
\end{table*}

In this section, we provide a brief description of the participating methods. Training and inference times for different methods are summarized in Table~\ref{tab:team_info}.

\subsection{Noah\_TerminalVision\_SR}
\begin{figure*}[t]
    \centering%
    \includegraphics[width=0.99\textwidth]{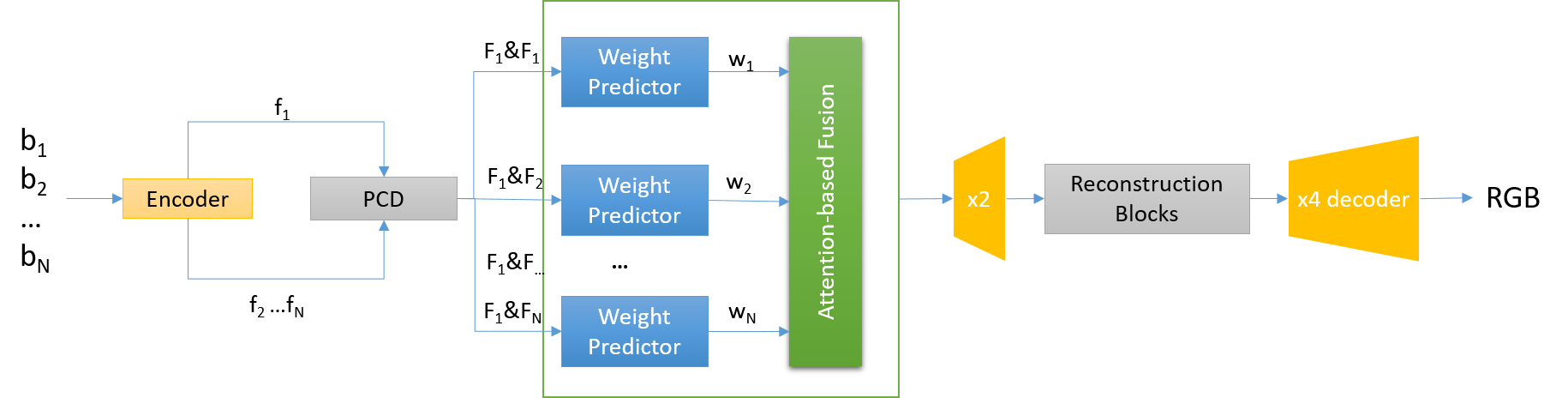}
    \caption{Overview of the NoahBurstSRNet network architecture employed by team Noah\_TerminalVision\_SR}\vspace{0mm}
    \label{fig:noah_network}
\end{figure*}

The team proposes NoahBurstSRNet for the task of Joint Demosaicking, Denoising and Super Resolution (JDDSR) of smart-phone burst Raw images. NoahBurstSRNet is inspired by two recent works: EDVR \cite{wang2019edvr} and DBSR~\cite{Bhat2021DeepBS}. The network architecture is shown in Figure~\ref{fig:noah_network}.

Burst Raw images are passed to an encoder to extract features. Those features are then aligned to the base feature (the feature of the first Raw image) by a Pyramid, Cascading and Deformable (PCD) \cite{wang2019edvr} module. With the aligned features, a Weight Predictor (WP) is used to predict the fusion weights between the base frame features and the other frame features. Based on the predicted fusion weights, an Attention-based Fusion (ABF) module is used to fuse the aligned futures. The WP and ABF modules are very similar to those proposed in~\cite{Bhat2021DeepBS}, except that the optical flow is not used as an input of WP. The spatial resolution of the fused features is first increased by a factor of two using a Pixelshuffle layer~\cite{Shi2016RealTimeSI}. The upsampled feature map is then processed by a sequence of Residual Feature Distillation Blocks (RFDBs) \cite{liu2020residual} in order to reconstruct the high-frequency details. Finally, the spatial resolution of the feature map is further increased by a factor of 4 and the final RGB image is predicted using pixelshuffle and convolution layers.

For track 1, the NoahBurstSRNet was trained in a fully-supervised manner on synthetic bursts generated using sRGB images from the Zurich RAW to RGB training set. The network was trained using the L1 loss. For the final submission, the team used a self-ensemble technique proposed in~\cite{liu2019learning} which can augment the input data while preserving the bayer pattern.

For track2, the team employed a two-stage training strategy. For the first stage, since the training data is weakly-paired (\ie they are not pixel-wise aligned), the model was trained using the spatial and color alignment strategy proposed in~\cite{Bhat2021DeepBS}. This model is termed as Noah\_TerminalVision\_SR\_A.
Next, the ground truth linear RGB images were spatially aligned and color matched to the SR predictions of the model trained in stage 1 using an in-house tool. The resulting aligned ground truth images were then used to fine-tune the model in stage 2 to obtain Noah\_TerminalVision\_SR\_B. The fine-tuning was performed using the SSIM and LPIPS \cite{Zhang2018TheUE} loss as these losses can tolerate small misalignments in the training data.

\subsection{MegSR}
\begin{figure*}[t]
    \centering%
    \includegraphics[width=0.99\textwidth]{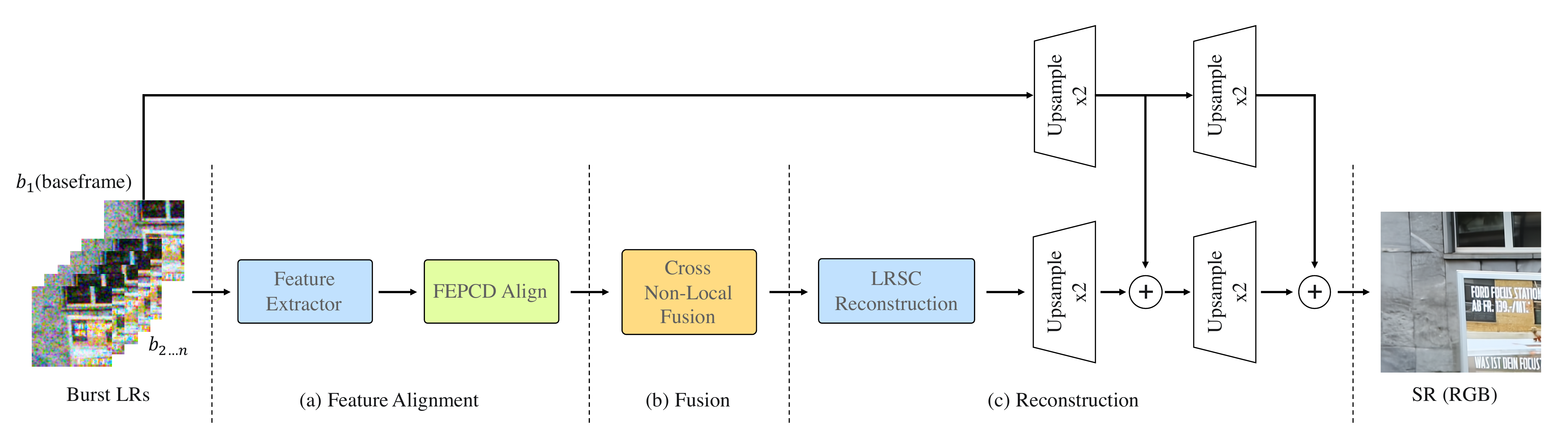}
    \caption{Overview of the network architecture employed by team MegSR.}\vspace{0mm}
    \label{fig:megsr_pipeline}
\end{figure*}

The team MegSR propose Feature Enhanced Burst Super-Resolution with Deformable Alignment (EBSR) framework, as shown in Fig.~\ref{fig:megsr_pipeline}.
EBSR solves the burst SR problem in three steps: align, fusion and reconstruction. First, it extracts high-level features of the LR burst images. The features are then aligned by a Feature Enhanced Pyramid, Cascading and Deformable convolution (FEPCD) module. Next, the aligned features are fused by a Cross Non-Local Fusion (CNLF) module. Finally, the SR image is reconstructed by the Long Range  Concatenation Network (LRCN). In addition, EBSR builds a progressively scaled residual pathway structure to further improve the performance. Please refer to~\cite{luo2021ebsr} for a detailed description.

\parsection{Feature Alignment}
EBSR extracts high level features using Wide activation Residual Block (WARB) introduced in~\cite{yu2018wide}. The features are aligned by a Feature Enhance PCD (FEPCD) module with multi-scale features. FEPCD is an extension of the PCD module~\cite{wang2019edvr} where an initial feature pyramid is used to first denoise and enhance the image features
The feature of the first frame from the LR image is chosen as the reference, and the features from other LR images are aligned to this reference.

\parsection{Fusion} 
In order to properly fuse the features from different frames, EBSR introduces Cross Non-Local (CNL) module which is based on Non-Local network~\cite{Wang_2018_CVPR}. The CNL fusion module measures the similarity between every two pixels from reference frame and other frame feature maps. The more similar the feature representations between two locations, higher the correlation between them. According to this property, the valid regions of other frames are fused into the reference frame. 

\parsection{Reconstruction}
The final SR output is reconstructed by the Long Range Concatenation Network (LRCN). It consists of two parts: backbone module and upsample module. The backbone module is composed of $G$ Long-Range Concatenation Groups (LRCG), and each LRCG contains $N$ Long-Range Residual Blocks with wide activation which is inspired by WDSR \cite{yu2018wide}.
Moreover, EBSR introduces a progressive upsample module with pixelShuffle~\cite{huang2009multi} and a residual pathway structure to reconstruct the final SR image. 
The reference frame goes through a pixelshuffle layer, and is added to the outputs of pixelshuffle layers of reconstruction module. 

\parsection{Training} 
For Track 1, the network is trained using the charbonnier loss proposed by LapSRN~\cite{lai2017deep}. The trained model is then fine-tuned on the BurstSR train set to obtain the model for Track 2. When training on BurstSR dataset, the network predictions are first aligned to the ground truth as described in~\cite{Bhat2021DeepBS}. 
In order to further improve the performance, the team employs a multi-model ensemble training strategy. Three trained EBSR models are loaded with frozen weights, and a few additional convolution layers are trained to fuse their outputs.

\parsection{Testing}
Team MegSR uses the Test Time Augmentation (TTA) strategy which can be seen as a self-ensemble approach. Specifically, given the original input burst, each image in the burst is transposed and the images in the burst are shuffled. The two augmented bursts, along with the original burst, are then passed through the network. The resulting outputs are averaged after reversing the augmentation effect to obtain the final prediction.

\subsection{Inria}
\begin{figure*}[t]
    \centering%
    \includegraphics[width=0.99\textwidth]{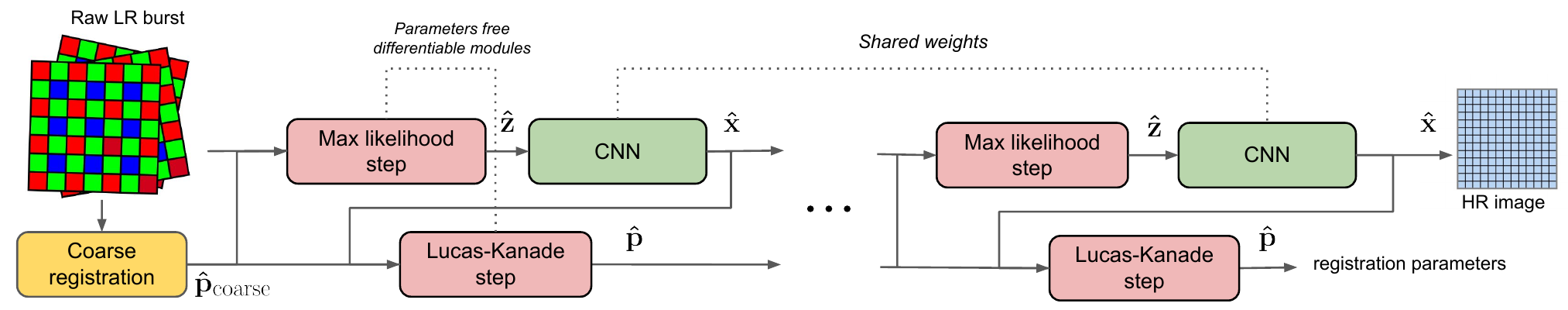}
    \caption{Overview of the network architecture employed by team Inria}\vspace{0mm}
    \label{fig:inria_pipeline}
\end{figure*}

\newcommand\vsp{\vspace*{-0.2cm}}
\newcommand\yb{\mathbf{y}}
\newcommand\ab{\mathbf{a}}
\newcommand\tmone{{t\text{--}1}}
\newcommand\Yb{\mathbf{Y}}
\newcommand\Ncal{\mathcal{N}}
\newcommand\vb{\mathbf{v}}
\newcommand\kb{\mathbf{k}}
\newcommand\xb{\mathbf{x}}
\newcommand\Xb{\mathbf{X}}
\newcommand\Hb{\mathbf{H}}
\newcommand\Db{\mathbf{D}}
\newcommand\Ab{\mathbf{A}}
\newcommand\Ub{\mathbf{U}}
\newcommand\Pb{\mathbf{P}}
\newcommand\Fb{\mathbf{F}}
\newcommand\Zb{\mathbf{Z}}
\newcommand\Jb{\mathbf{J}}
\newcommand\ub{\mathbf{u}}
\newcommand\pb{\mathbf{p}}
\newcommand\wb{\mathbf{w}}
\newcommand\zb{\mathbf{z}}
\newcommand\Cb{\mathbf{C}}
\newcommand\Ib{\mathbf{I}}
\newcommand\Wb{\mathbf{W}}
\newcommand\Rb{\mathbf{R}}
\newcommand\Mb{\mathbf{M}}
\newcommand\Bb{\mathbf{B}}
\newcommand\db{\mathbf{d}}
\newcommand\bb{\mathbf{b}}

\newcommand\rb{\mathbf{r}}
\newcommand\varepsilonb{{\boldsymbol \varepsilon}}

\newcommand\Tb{\mathbf{T}}

\newcommand\Fcal{\mathcal{F}}

\newcommand\ones{\mathbf{1}}
\newcommand\Real{\mathbb{R}}
\newcommand\alphab{\mathbf{z}}
\newcommand\betab{\boldsymbol{\beta}}
\newcommand\Lambdab{\boldsymbol{\Lambda}}
\newcommand\Thetab{\boldsymbol{\Theta}}
\newcommand\thetab{\boldsymbol{\theta}}
\newcommand\Sigmab{\boldsymbol{\Sigma}}
\newcommand\kappab{\boldsymbol{\kappa}}
\newcommand\lambdab{\boldsymbol{\lambda}}

\newcommand\Pcal{{\mathcal{P}}}

\newcommand{\norm}[1]{\lVert#1\rVert}

\def\eg{\emph{e.g.}} 

\newcommand{\blue}[1]{\textcolor{blue}{#1}}

\def\texte#1{#1}

\def\RR{\mathbb{R}}
\def\CC{\mathbb{C}}
\def\KK{\mathbb{K}}
\def\NN{\mathbb{N}}
\def\PP{\mathbb{P}}
\def\AA{\mathbb{A}}
\def\LL{\mathbb{L}}
\def\SS{\mathbb{S}}
\def\barr{\bar{\mathbb{R}}}
\def\mat#1{{\mathcal{#1}}}
\def\vect#1{\mbox{\boldmath $#1$}}
\def\PPi{\mbox{\boldmath$\Pi$}}
\def\squig{\rightsquigarrow}
\def\eqdef{\buildrel \rm def \over =}

\def\comment#1{{}}
\def\qmatrix#1{\left[\begin{matrix}#1\end{matrix}\right]}
\def\st#1{{\tt #1}}
The method presented in this section corresponds to the paper
\cite{Lecouat2021AliasingIY}, where all details can be found. The brief description next, including figures, is borrowed from this paper.
Starting from a low resolution burst of raw images, team Inria 
estimates a coarse block-parametric displacement
field using a robust multiscale Lucas-Kanade
algorithm~\cite{baker2004lucas}. Next, an estimate
of the high resolution image is obtained and the motion parameters are subsequently refined alternatively.
This iterative process involves a data-driven
prior---here, a convolutional neural network---which helps
removing artefacts. Model parameters are learned end-to-end by
backpropagating on real HR/synthetic LR examples. An overview of the network architecture is provided in Figure~\ref{fig:inria_pipeline}. The HR estimate as well as the refined motion parameters are obtained using the image formation model described next.

\parsection{Inverse problem and optimization}
The burst images are obtained through
the following forward model:
\begin{equation}
  \yb_k=DBW_{\pb_k}\, \xb+\varepsilonb_k \,\,\text{for}\,\,k=1,\ldots,K, 
  \label{eq:imfor}
\end{equation}
where $\varepsilonb_k$ is additive noise. Here, both the HR
image~$\xb$ and the frames $\yb_k$ of the burst are flattened into
vector form. The operator $W_{\pb_k}$, parameterized by
$\pb_k$, warps $\xb$ to compensate for misalignments between $\xb$ and
$\yb_k$ caused by camera or scene motion between frames and
resamples the warped image to align its pixel grid with that of
$\yb_k$. Finally, the corresponding HR image is blurred to account for
integration over space, and it is finally downsampled in
both the spatial and spectral domains by the operator $D$. The spectral part corresponds to
mosaicking operation of selecting one of the three RGB values to assemble the raw image. The model~\eqref{eq:imfor} can be rewritten as
$\yb=U_\pb\xb+\varepsilonb$, where
\begin{equation}
  U_{\pb}\!=\!\qmatrix{DBW_{\pb_1}\\ \vdots\\ DBW_{\pb_K}}\!\!,
  \yb\!=\!\qmatrix{y_1\\ \vdots\\ y_K}\!\!,\pb=\qmatrix{p_1\\ \vdots
    \\p_K}\!\!,
  \varepsilonb\!=\!\qmatrix{\varepsilon_1\\ \vdots\\ \varepsilon_K}\!.
  \label{eq:imforc}
\end{equation}

Given the image formation model of
Eq.~\eqref{eq:imfor}, recovering the HR image $\xb$ from the $K$ LR frames $\yb_k$ in the burst
can be formulated as finding the values of $\xb$ and $\pb$ that minimize
\begin{equation}
 \frac{1}{2}\|\yb-U_\pb\, \xb\|^2+\lambda\phi_\theta(\xb),
  \label{eq:optprob}
\end{equation}
where $\phi_\theta$ is a parameterized regularizer, which will be chosen as a convolutional neural network, and $\lambda$ is a parameter  balancing the
data-fidelity and regularization terms. The objective~\eqref{eq:optprob} is minimized using the
quadratic penalty method~\cite[Sec.~17.1]{nocedal} often called
half-quadratic splitting (or HQS)~\cite{geman1995nonlinear}. Here, the
original objective is replaced by
\begin{equation}
  E_\mu(\xb,\zb,\pb)= \frac{1}{2}\|\yb-U_\pb\,
  \zb\|^2+\frac{\mu}{2}\|\zb-\xb\|^2
  +\lambda \phi_\theta(\xb),
  \label{eq:HQS}
\end{equation}
where $\zb$ is an auxiliary variable, and $\mu$ is a parameter
increasing at each iteration, such that, as $\mu\rightarrow+\infty$,
the minimization of (\ref{eq:HQS}) with respect to $\xb$, $\zb$ and
$\pb$ becomes equivalent to that of (\ref{eq:optprob}) with respect to
$\xb$ and $\pb$ alone. The sequence
of weights $(\mu^{t})_{t \geq 0}$ are learned end-to-end.

\parsection{Estimating the HR image $\xb$}
The estimate $\xb$ is updated as
\begin{displaymath}
   \xb^t \leftarrow \mathrm{argmin}_{\xb} \frac{\mu_{\tmone}}{2}\|\zb^t-\xb\|^2+\lambda\phi_\theta(\xb),
\end{displaymath}
which amounts to computing the proximal operator of the prior
$\phi_\theta$.  Following the ``plug-and-play''
approach~\cite{chan2016plug,ryu2019plug,venkatakrishnan2013plug},  the proximal operator is replaced by a parametric function
$f_\theta(\zb_t)$ (here, a CNN). 

\parsection{Initialization by coarse alignement}
Each LR frame is aligned to an arbitrary one from the burst (\eg, the first one)
by using the robust Lucas-Kanade forward additive algorithm \cite{baker2004lucas,sanchez2016inverse} which is known to be robust to noise.
The raw images are first converted to grayscale format by
using bilinear interpolation. Although sub-optimal, such a procedure is sufficient
for obtaining coarse motion parameters. 

\parsection{Learned data prior}
Good image priors are essential for solving ill-posed inverse
problems.  Instead of using a classical one, such
as total variation (TV), team Inria learns an implicit prior parameterized
by a convolutional neural network $f_\theta$ in a data-driven
manner. The ResUNet architecture proposed in~\cite{zhang2020deep} is employed for this purpose. 

\parsection{Training} The model is trained using synthetic bursts generating using sRGB images from the
training split of the Zurich raw to RGB dataset~\cite{ignatov2020replacing}, by minimizing the $\ell_1$ loss.

\subsection{TTI}

\begin{figure*}[t]
    \centering
    \includegraphics[width=0.9\textwidth]{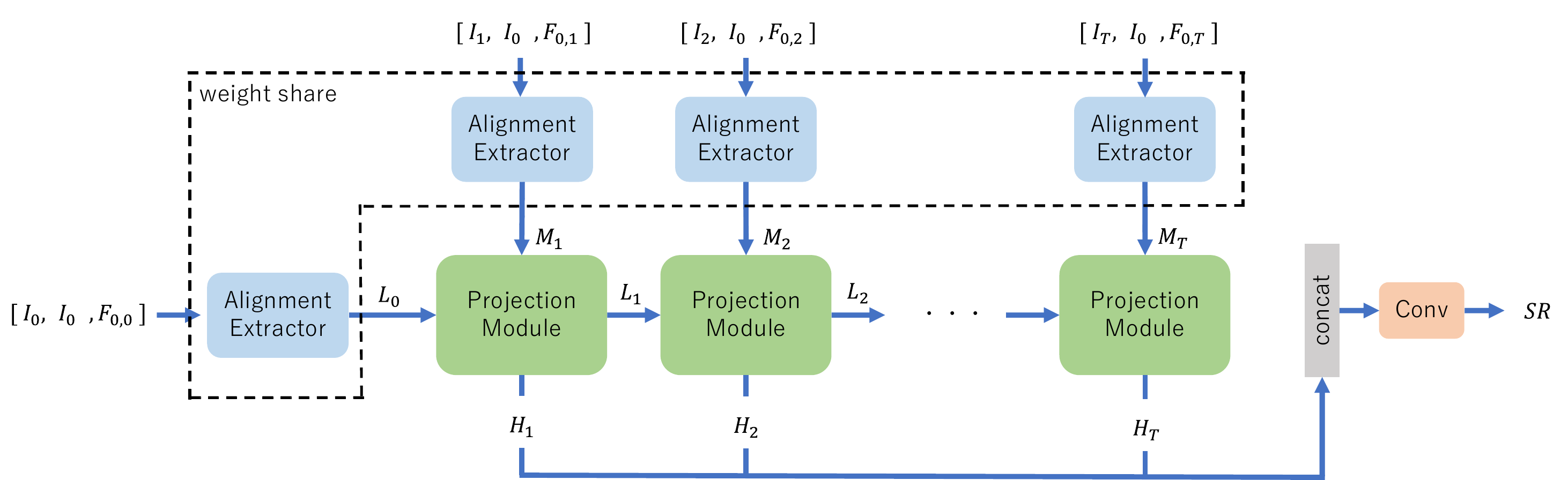}
    \caption{Overview of the network architecture employed by team TTI.}
    \label{fig:tti_overview}
\end{figure*}

The team TTI adopt the Recurrent Back Projection Network (RBPN)~\cite{DBLP:conf/cvpr/HarisSU19} for super-resolving burst frames. 
RBPN constructs the projection module inspired by DBPN~\cite{DBLP:conf/cvpr/HarisSU18, DBLP:conf/cvpr/HarisSU19} to iteratively refine an input frame using temporally-nearby frames, along with corresponding flow vectors. An overview of the approach is provided in Figure~\ref{fig:tti_overview}.
In addition to the basic functions in RBPN, TTI applies deformable convolutions to align features to the input frame.
A flow refinement module is also proposed to minimize the flow estimation error caused by noisy frames.
This module is trained explicitly using the ground-truth flow in track 1 but trained implicitly solely on SR reconstruction loss in track 2. 
The residual blocks in the projection module are initialized using Fixup Initialization~\cite{DBLP:conf/iclr/ZhangDM19} to stabilize the training process of the normalization-free network.

\parsection{Flow Estimation and Refinement} 
For initial flow estimation between a reference frame $I_0$ and frame $I_t$, TTI utilizes PWCNet \cite{Sun2018PWCNetCF} with fixed pretrained weights.
In order to further refine the predicted flow, 
a UNet-like flow refinement module is employed.
The estimated flow is concatenated with the reference frame $I_0$ and frame $I_t$, and fed into the refinement module to get the residual flow, which is then added to the initial flow estimate.

\parsection{Alignment Extractor}
In the original RBPN, a single convolution layer is used for feature extraction from each input frame, neighboring frame and the corresponding optical flow triplet. 
In contrast, team TTI first concatenates features from reference frame $I_0$ and frame $I_t$, along with the corresponding optical flow $F_{0, t}$ and passes this through a convolution layer to obtain offsets. These offsets are then used to align the frame $I_t$ features to the base frame using a deformable convolution layer.

\parsection{Training}
For Track 1, the model is trained on synthetic burst data generated using sRGB image from~\cite{ignatov2020replacing}, using the $L_1$ loss. Additionally, $L_1$ loss between the ground-truth flow $\hat{F}$ and refined flow $F$ is also used to train the flow refinement module.
In Track 2, the model was trained on BurstSR training set, using the spatial and color alignment strategy employed in~\cite{Bhat2021DeepBS}, by minimizing the $L_2$ loss. Since ground-truth flow $\hat{F}$ is not available in this case, the flow refinement module was trained using only the final SR reconstruction loss, without any direct supervision.

\parsection{Inference}
To achieve a fast training process and memory efficiency, the model is trained using bursts containing 8 images. During inference, original burst containing 14 frames is divided into 7 subsets, each including the reference (first) frame. The 7 SR predictions are averaged to output the final SR image. No model-ensemble and self-ensemble are used.

\subsection{MLP\_BSR}
\begin{figure*}[t]
    \centering
    \includegraphics[width=0.9\textwidth]{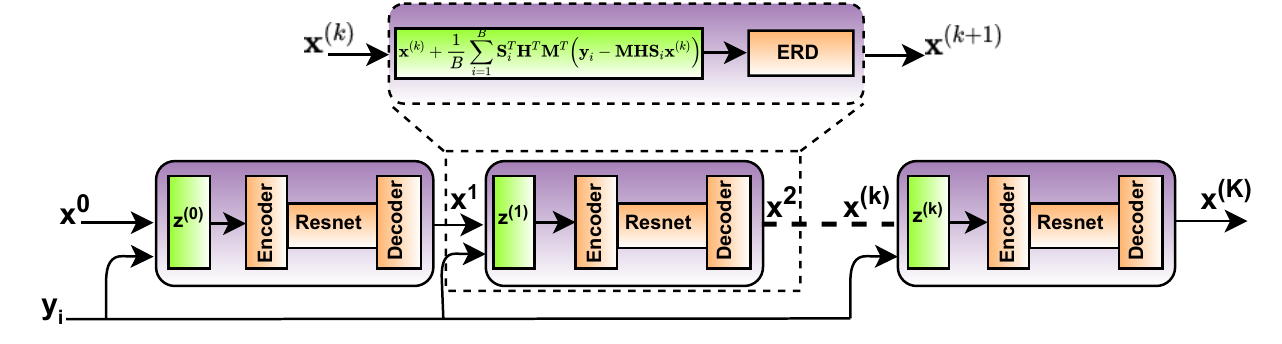}
    \caption{The overview the architecture employed by team MLP\_BSR.}
    \label{fig:bsricnn}
\end{figure*}

\def\ba{{\bf a}}
\def\bb{{\bf b}}
\def\bc{{\bf c}}
\def\bd{{\bf d}}
\def\be{{\bf e}}
\def\bg{{\bf g}}
\def\bh{{\bf h}}
\def\bk{{\bf k}}
\def\bl{{\bf l}}
\def\bn{{\bf n}}
\def\bs{{\bf s}}
\def\bu{{\bf u}}
\def\bv{{\bf v}}
\def\bw{{\bf w}}
\def\bx{{\bf x}}
\def\by{{\bf y}}
\def\bz{{\bf z}}

\def\0{{\bf 0}}
\def\1{{\bf 1}}

\def\bA{{\bf A}}
\def\bB{{\bf B}}
\def\bC{{\bf C}}
\def\bD{{\bf D}}
\def\bE{{\bf E}}
\def\bG{{\bf G}}
\def\bH{{\bf H}}
\def\bI{{\bf I}}
\def\bM{{\bf M}}
\def\bN{{\bf N}}
\def\bP{{\bf P}}
\def\bS{{\bf S}}
\def\bZ{{\bf Z}}
\def\bR{{\bf R}}
\def\bK{{\bf K}}
\def\bX{{\bf X}}
\def\bY{{\bf Y}}
\def\bZ{{\bf Z}}

\def\bmu{\boldmath{\mu}}
\def\blambda{\boldmath{\lambda}}
\def\balpha{\mbox{{\boldmath $\alpha$}}}
\def\bpsi {\mbox{{\boldmath $\psi$}}}
\def\bPsi{\mbox{{\boldmath $\Psi$}}}
\def\bPi{\mbox{{\boldmath $\Pi$}}}
\def\bphi{\mbox{{\boldmath $\phi$}}}
\def\bPhi{\mbox{{\boldmath $\Phi$}}}
\def\bUpsilon{\mbox{{\boldmath $\Upsilon$}}}
\def\bLambda{\mbox{{\boldmath $\Lambda$}}}
\def\boldeta{\mbox{{\boldmath $\eta$}}}
\def\bOmega{\mbox{{\boldmath $\Omega$}}}
\def\bomega{\mbox{{\boldmath $\omega$}}}
\def\bzeta {\mbox{{\boldmath $\zeta$}}}
\def\bvarpi{\mbox{{\boldmath $\varpi$}}}
\def\btau {\boldsymbol{\tau}}
\def\bbeta{\mbox{{\boldmath $\beta$}}}
\def\bDelta{\mbox{{\boldmath $\Delta$}}}
\def\bdelta{\mbox{{\boldmath $\delta$}}}
\def\bxi{\mbox{{\boldmath $\xi$}}}

\def\cA{{\mathcal A}}
\def\cC{{\mathcal C}}
\def\cD{{\mathcal D}}
\def\cF{{\mathcal F}}
\def\cG{{\mathcal G}}
\def\cH{{\mathcal H}}
\def\cK{{\mathcal K}}
\def\cL{{\mathcal L}}
\def\cN{{\mathcal N}}
\def\cM{{\mathcal M}}
\def\cS{{\mathcal S}}
\def\cR{{\mathcal R}}
\def\cT{{\mathcal T}}
\def\cU{{\mathcal U}}
\def\cW{{\mathcal W}}
\def\cX{{\mathcal X}}
\def\cY{{\mathcal Y}}
\def\cZ{{\mathcal Z}}

\def\mbA{{\mathbb A}}
\def\mbN{{\mathbb N}}
\def\mbR{{\mathbb R}}

\def\tx{\tilde{\x}}
\def\barx{\bar{\x}}
\def\tbPi{\mbox{{\boldmath $\tilde{\Pi}$}}}

\def\diag{\mbox{diag}}
\def\dist{\mbox{dist}}
\def\sgn{\mbox{sgn}}
\def\card{{\mbox{Card}}}

\def\etal{\emph{et al. }}
\def\ie{\emph{i.e. }}
\def\eg{\emph{e.g. }}
\def\etc{\emph{etc }}
\def\supp{\text{supp}}
\def\st{{\mathrm{s.t.}}}

\def\la{\langle}
\def\ra{\rangle}

\def\T{{\sf T}}

\def\DIFFMAP{\mbox{{ ${\Phi_{\y_i,\y}}$}}}
\def\DIFFMAPCL{\mbox{{ ${\Phi^\mC_{\y_i,\y}}$}}}
\def\DIFFMAPCLK{\mbox{{ ${\Phi^{\mC^t}_{\y_i,\y}}$}}}
\def\DIFFMAPND{\mbox{{ ${\Phi^\mV_{\y_i,\y}}$}}}
\def\DIFFMAPcl{\mbox{{ ${\Phi^\mc_{\y_i,\y}}$}}}
\def\DIFFMAPCLKHAT{\mbox{{ ${\Phi^{\mU^t}_{\y_i,\y}}$}}}
\def\DIFFMAPGAMMA{\mbox{{ ${\Phi^{\Gamma_k}_{\y_i,\y}}$}}}
\def\DIFFMAPU{\mbox{{ ${\Phi^{\mU}_{\y_i,\y}}$}}}

\def\grad{{\nabla}}

\def\taux{\bm{\tau}_{\bf x}}
\def\kappax{\kappa_{\bf x}}
\def\kappaw{\kappa_{\bf w}}
\def\Lambdax{\Lambda_{\bf x}}
\def\Lambdaw{\Lambda_{\bf w}}
\def\bxi{\bm{\xi}}

\newcommand\mynote[1]{\textcolor{RedOrange}{#1}}

The team MLP\_BSR propose a deep iterative Burst SR learning method (BSRICNN) that solves the Burst SR task in an iterative manner. To solve the Raw Burst Super-Resolution task, they rely on the forward observation model,
\begin{equation}
    \by_i = \bM \bH \bS_i (\Tilde{\bx}) + \eta_i, ~~~ i = 1,\ldots, B
    \label{eq:degradation_model}
\end{equation}
where, $\by_i$ is an observed LR burst containing $B$ images, $\bM$ is a mosaicking operator that corresponds to the CFA (Color Filter Array) of a camera (usually Bayer), $\bH$ is a down-sampling operator (\ie bilinear, bicubic, etc.) that resizes an HR image $\Tilde{\bx}$ by a scaling factor $r$, $\bS_i$ is an affine transformation of the coordinate system of the image $\Tilde{\bx}$ (\ie translation and rotation), and $\eta_i$ is an additive heteroscedastic Gaussian noise related to shot and read noise. Due to the ill-posed nature of inverse problem, the recovery of $\bx$ from $\by_i$ mostly relies on variational approaches for combining the observation and prior knowledge, and the solution is obtained by minimizing the following objective function as,
\begin{equation}
    \hat{\bx} = \underset{\mathbf{x}}{\arg \min }~\frac{1}{2\sigma^2B}\sum_{i=1}^{B}\|\by_i - \bM \bH \bS_i(\bx)\|_2^{2}+\lambda \mathcal{R}(\bx),
    \label{eq:eq1}
\end{equation}
where, the first term is a data fidelity term that measures the proximity of the solution to the observations, the second term (\ie $\mathcal{R}(\bx)$) is the regularization term that is associated with image priors, and $\lambda$ is the trade-off parameter that governs the compromise between the data fidelity and the regularizer term. The objective~\eqref{eq:eq1} is minimized using Majorization-Minimization (MM) framework~\cite{hunter2004tutorial} which has been previously employed in image restoration tasks~\cite{Umer_2020_ICPR,Kokkinos2019IterativeJI}.

The proposed Burst SR scheme is shown in Figure~\ref{fig:bsricnn}.  BSRICNN is unrolled into $K$ stages, where each stage computes the refined estimate of the solution. $\by_i$ is an input Raw LR burst, $\bx^0$ is an initial estimate, and $\bx^K$ is a final estimated SR image. In Figure~\ref{fig:bsricnn}, the Encoder-Resnet-Decoder (ERD) architecture employed is similar to the one used in~\cite{Umer_2020_ICPR}. Each network stage performs the efficient Proximal Gradient Descent~\cite{ParikhPGM} updates to solve the optimization problem.
The shared parameters across stages are learned jointly by minimizing the $L_1$ loss \wrt to all network parameters. For Track 1, the network is trained using synthetic bursts generated using 46,839  sRGB images from the Zurich RAW to RGB dataset~\cite{ignatov2020replacing}. The same model was then also employed for Track 2, without any finetuning on the BurstSR dataset.

\subsection{BREIL}
\begin{figure*}[t]
    \centering
    \includegraphics[width=0.9\textwidth]{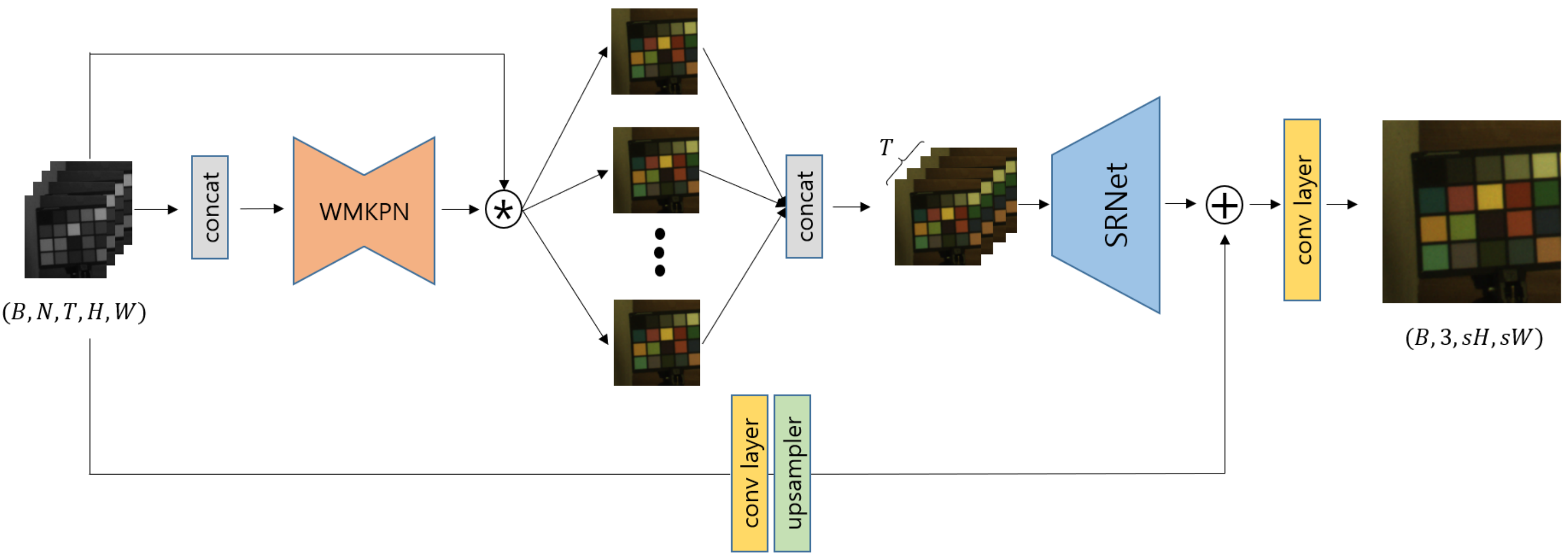}
    \caption{Overview of the network architecture employed by team BREIL.}
    \label{fig:breil}
\end{figure*}

The proposed model takes multiple LR RAW burst images with noise $\left\{b_{i}\right\}^T_{i=1}$ and predicts denoised single HR RGB image $I^i_{HR}$. The kernel prediction network (KPN)~\cite{mildenhall2018burst,marinvc2019multi} which has been recently employed for burst image processing is utilized for aligning the images from the burst. The overall model shown in Figure~\ref{fig:breil} consists of two parts, i) a KPN-based module that aligns the burst images and ii) a reconstruction network that increases the resolution of images while fusing the output of the alignment module. Each of these modules are briefly described next. Please refer to~\cite{cho2021wmkpn} for more details.

\parsection{Weighted Multi-Kernel Prediction}
The team BREIL propose a weighted multi-kernel prediction network (WMKPN) for aligning the images from the input burst. WMKPN is an extension of MPKN approach introduced in~\cite{marinvc2019multi}. MKPN predicts multiple kernels of different sizes, which are then used to align and fuse the burst images. All the kernels are weighted equally in the MKPN framework. In contrast, WMKPN predicts additional weights for each kernel to adaptively weight the impact different kernels. Furthermore, the predicted kernels are only used to align the input images, while the fusion is performed by a seperate network.
WMKPN employs a modified U-net architecture, resembling \cite{mildenhall2018burst,marinvc2019multi,zhang2020attention}. In the encoder part, the spatial sizes of the feature maps are reduced by the average pooling layer while using the convolution layer and ReLU activation function. On the other side, the decoder increases the spatial sizes of the feature maps by a bilinear upsampling layer. WMKPN also exploits the convolution layer and attention module, proposed in \cite{zhang2020attention}, which is composed of a series of the channel attention (CA) \cite{zhang2018image} and the spatial attention (SA) \cite{hu2019spatial}. The encoded feature maps are concatenated to the decoder side that have the same spatial sizes like U-net architecture \cite{unet}. This modified U-net consists of two branches, the kernel prediction branch and the kernel weight branch which predict kernels and weights, respectively. Separable kernels are used for memory efficiency. To provide discriminative mechanism, the kernel weights are normalized using softmax operation applied across the different kernels.  
The weighted kernels are then convolved with the input burst images in order to align them.

\parsection{SR Reconstruction Network}
The aligned input images are concatenated and passed through a reconstruction network to obtain the SR image.
The reconstruction network leverages the residual blocks with short connections which are exploited in EDSR approach~\cite{lim2017enhanced}. The SR network extracts deep features by passing the input through ten residual blocks. The output is then upsampled using three sub-pixel convolution layers~\cite{Shi2016RealTimeSI} to obtain the final RGB image.

\parsection{Training} The network is trained using a combination of $L_1$ and SSIM losses.

\section{Conclusion}
This paper describes the NTIRE2021 challenge on burst super-resolution. Given multiple images of a scene captured in quick succession, the burst super-resolution task aims to generate a super-resolved output by merging information from the input frames. The challenge tackled the problem of RAW burst super-resolution, where the goal is to predict a 4x super-resolved RGB image, given a RAW noisy burst as input. The challenge contained two tracks, namely Track 1 and Track 2. In Track 1, a synthetically generated burst dataset was used for evaluation, while Track 2 focused on real-world SR using bursts captured from a hand held camera. 6 teams submitted results in the final testing phase. The participating methods, described in this report, employed a diverse set of approach for the burst SR problem obtaining promising results.

\noindent\textbf{Acknowledgments}: We thank the NTIRE 2021 sponsors: Huawei, Facebook Reality Labs, Wright Brothers Institute, MediaTek, OPPO and ETH Zurich (Computer Vision Lab). We also thank Andreas Lugmayr for helping with the human study.

\setcounter{section}{0}

\renewcommand{\thesection}{\Alph{section}}

\begin{center}
	\textbf{\large Appendix}
\end{center}

\section{Teams and Affiliations}

\subsection*{NTIRE2021 Organizers}
\parsection{Members} \\
Goutam Bhat (goutam.bhat@vision.ee.ethz.ch) \\
Martin Danelljan (martin.danelljan@vision.ee.ethz.ch) \\
Radu Timofte (radu.timofte@vision.ee.ethz.ch) \\
\parsection{Affiliation} Computer Vision Lab, ETH Zurich

\subsection*{Noah\_TerminalVision\_SR}
\parsection{Title} Real-World Joint Demosaicking, Denoising and Super Resolution of Smart-phone Burst Raw Images \\
\parsection{Team Leader} \\
Xueyi Zou (zouxueyi@huawei.com) \\
\parsection{Members} \\
Xueyi, Zou, Noah's Ark Lab, Huawei \\
Magauiya, Zhussip, Noah's Ark Lab, Huawei \\
Pavel, Ostyakov, Noah's Ark Lab, Huawei \\
Ziluan, Liu, CBG AITA, Huawei \\
Youliang, Yan, Noah's Ark Lab, Huawei \\

\subsection*{MegSR
}
\parsection{Title} EBSR: Feature Enhanced Burst Super-Resolution with Deformable Alignment
 \\
\parsection{Team Leader} \\
Lanpeng Jia
 (jialanpeng@megvii.com) \\
\parsection{Members} \\
Ziwei Luo, Megvii Technology \\
Lei Yu, Megvii Technology \\
Xuan Mo, Megvii Technology \\
Youwei Li, Megvii Technology \\
Lanpeng Jia, Megvii Technology \\
Haoqiang Fan, Megvii Technology \\
Jian Sun, Megvii Technology \\
Shuaicheng Liu, Megvii Technology \\

\subsection*{Inria
}
\parsection{Title} End-to-End Super-Resolution from Raw Image Bursts
 \\
\parsection{Team Leader} \\
Bruno Lecouat
 (bruno.lecouat@inria.fr) \\
\parsection{Members} \\
Bruno Lecouat, Inria \\
Jean Ponce, Inria \\
Julien Mairal, Inria \\

\subsection*{TTI}
\parsection{Title} Recurrent Back Projection Network for Burst Super-Resolution \\
\parsection{Team Leader} \\
Takahiro Maeda (sd19445@toyota-ti.ac.jp) \\
\parsection{Members} \\
Takahiro Maeda, Toyota Technological Institute \\
Kazutoshi Akita, Toyota Technological Institute \\
Takeru Oba, Toyota Technological Institute \\
Norimichi Ukita, Toyota Technological Institute \\

\subsection*{MLP\_BSR
}
\parsection{Title} Deep Iterative Convolutional Neural Network for Raw Burst Super-Resolution \\
\parsection{Team Leader} \\
Rao Muhammad Umer
 (engr.raoumer943@gmail.com) \\
\parsection{Members} \\
Rao Muhammad Umer, University of Udine, Italy \\
Christian Micheloni, University of Udine, Italy \\

\subsection*{BREIL
}
\parsection{Title} Weighted Multi-Kernel Prediction Network for Burst Image Super-resolution
 \\
\parsection{Team Leader} \\
Wooyeong Cho
 (chowy333@kaist.ac.kr) \\
\parsection{Members} \\
Wooyeong, Cho, KAIST (Korea Advanced Institute of Science and Technology) \\
Sanghyeok, Son, KAIST (Korea Advanced Institute of Science and Technology) \\
Daeshik, Kim, KAIST (Korea Advanced Institute of Science and Technology) \\

{\small
\bibliographystyle{ieee_fullname}
\bibliography{references}
}
\end{document}